\documentclass[]{elsarticle}
\usepackage[T1]{fontenc}
\usepackage{float}
\usepackage{graphicx}
\usepackage{booktabs}
\usepackage[numbers]{natbib}
\usepackage{amsmath}
\usepackage{graphicx}
\usepackage{multirow}
\usepackage{booktabs}
\usepackage{rotating} 
\usepackage{makecell} 
\usepackage{tabularx}
\usepackage{hyperref}

\begin{document}
\begin{frontmatter}
\title{From Cold Start to Active Learning: Embedding-Based Scan Selection for Medical Image Segmentation}

% Authors
\author[inst1]{Devon Levy}
\author[inst1]{Bar Assayag}
\author[inst2]{Laura Gaspar}
\author[inst1]{Ilan Shimshoni}
\author[inst2]{Bella Specktor-Fadida\corref{cor1}}
\ead{bspecktor@univ.haifa.ac.il}

% Affiliations
\affiliation[inst1]{organization={Department of Information Systems, University of Haifa},
  country={Israel}}

\affiliation[inst2]{organization={Department of Medical Imaging Sciences, University of Haifa},
  country={Israel}}

\cortext[cor1]{Corresponding author.}

\begin{abstract}
\noindent 
Accurate segmentation annotations are critical for disease monitoring, yet manual labeling remains a major bottleneck due to the time and expertise required. Active learning (AL) alleviates this burden by prioritizing informative samples for annotation, typically through a diversity-based cold-start phase followed by uncertainty-driven selection. We propose a novel cold-start sampling strategy that combines foundation-model embeddings with clustering, including automatic selection of the number of clusters and proportional sampling across clusters, to construct a diverse and representative initial training. This is followed by an uncertainty-based AL framework that integrates spatial diversity to guide sample selection. The proposed method is intuitive and interpretable, enabling visualization of the feature-space distribution of candidate samples. We evaluate our approach on three datasets spanning X-ray and MRI modalities. On the CheXmask dataset, the cold-start strategy outperforms random selection, improving Dice from 0.918 to 0.929 and reducing the Hausdorff distance from 32.41 to 27.66 mm. In the AL setting, combined entropy and diversity selection improves Dice from 0.919 to 0.939 and reduces the Hausdorff distance from 30.10 to 19.16 mm. On the Montgomery dataset, cold-start gains are substantial, with Dice improving from 0.928 to 0.950 and Hausdorff distance decreasing from 14.22 to 9.38 mm. On the SynthStrip dataset, cold-start selection slightly affects Dice but reduces the Hausdorff distance from 9.43 to 8.69 mm, while active learning improves Dice from 0.816 to 0.826 and reduces the Hausdorff distance from 7.76 to 6.38 mm. Overall, the proposed framework consistently outperforms baseline methods in low-data regimes, improving segmentation accuracy.

\end{abstract}

%\begin{highlights}
%\item We propose a novel cold-start sample selection method for segmentation that combines foundation-model embeddings with clustering, automatic determination of the number of clusters and proportional sampling.
%\item We introduce an active learning strategy that combines uncertainty and spatial diversity to guide sample selection.
%\item The proposed framework operates in a low-dimensional embedding space, enabling natural explainability of the selection process while maintaining low computational overhead.
%\item Experiments on three datasets across X-ray and MRI modalities and different anatomical structures show consistent improvements over random selection and a competitive baseline.
%\end{highlights}

\begin{keyword}
Active learning, Segmentation, Neural networks, Medical image analysis
%% keywords here, in the form: keyword \sep keyword

%% PACS codes here, in the form: \PACS code \sep code

%% MSC codes here, in the form: \MSC code \sep code
%% or \MSC[2008] code \sep code (2000 is the default)

\end{keyword}

\end{frontmatter}

\section{Introduction}
Segmenting anatomical structures in medical images is a key step in numerous clinical applications, including the evaluation of structures size and disease progression. Recent advances in deep learning have substantially improved segmentation performance across modalities \citep{litjens2017survey,rayed2024survey}. Although strong models can sometimes be trained with relatively small labeled datasets \citep{huysentruyt2024sample}, producing high-quality annotations remains expensive and time-consuming, making label-efficient learning strategies essential.

Active learning (AL) reduces annotation costs by iteratively selecting informative samples for labeling \citep{yang2017suggestive}. This is usually done by training a cold start initial network using diversity-oriented selection followed by uncertainty and sometimes in combination with a diversity method. For instance, Yuan et al. \cite{yuan2025rethinking} show that clustering foundation model embeddings yields a strong cold-start initializer for an initial network, followed by uncertainty-based acquisitions.
A pivotal stage is the cold start selection, where an initial batch is chosen when there are no labels. The performance of networks trained on randomly selected samples can vary considerably, and prior work has shown that deliberate sample selection usually leads to improved segmentation outcomes \citep{budd2021survey}. Early work examined clustering directly in pixel space. However, beyond very simple image datasets, raw pixel embeddings are poor proxies for semantics, and such baselines often perform poorly \citep{xie2016dec}.

Recent work selects samples using feature embeddings, either extracted via a foundation model or self-supervised pretraining \citep{zhou2023self}. Diversity-oriented selection in a learned representation space, such as farthest-first, where queries are chosen iteratively to maximize their minimum distance to the current labeled set, was also shown to be effective \citep{geifman2017ffactive}. Benchmark studies emphasize the importance of density-aware strategies in low-budget active learning regimes \citep{liu2023colossal}. The ALPS method applies k-means clustering to latent representations, with the number of clusters set equal to the query budget. CALR \citep{jin2022cold} employs a bottom-up hierarchical clustering approach, while TypiClust \citep{hacohen2022typiclust} first performs k-means clustering and then selects the most representative (“typical”) samples from each cluster. A comparative study by Liu et al. \cite{liu2023colossal} showed that among these approaches, only TypiClust consistently outperformed random selection. 

After an initial model training, pipelines that improve the initial network were proposed. They usually pair uncertainty with diversity to reduce redundancy. This may mean sampling uniformly from most uncertain cases to spread coverage \citep{rokach2025deepal}, or reasoning about uncertainty at the batch level to avoid near-duplicates \citep{gaillochet2023stochasticbatches}. In this work, we combine an image-level uncertainty score with feature-space diversity to favor informative yet non-redundant acquisitions.

Prior cold-start initializers often fix the number of clusters and then pick simple representative examples of either centroids, the samples closest to the mean of the cluster, or medoids, the points minimizing the total distance to others. With unknown class structure and heavy-tailed modes, an arbitrary \(K\) can collapse many samples into a few large clusters or over-select outliers and leave parts of the manifold under-covered, while purely uncertainty-driven rounds tend to re-query similar cases. We propose a unified framework for both the cold start and the subsequent active learning stage that automatically picks the optimal number of clusters for optimal coverage.

To address the challenges of cold-start annotation, in this work we propose a two-stage active learning framework that balances uncertainty and diversity in an intuitive and efficient way. The method first identifies a representative and diverse seed set that captures the major modes of the data distribution, then iteratively expands it by selecting samples that are both uncertain and distinct from previously labeled examples. By operating in a low-dimensional feature projection, our approach enables visual interpretation of the sampling process.

\section{Related Work}
Maximizing diversity within the training set has been a central strategy to improve sample efficiency across medical and non-medical domains. Diversity is typically employed in cold-start settings, where no trained model is yet available to guide sample selection. Geifman et al. \citep{geifman2017ffactive} select points by farthest-first traversal in embedding space to construct a diverse core set that scales efficiently across AL rounds. Subsequent works extend this idea through clustering and self-supervision: Chen et al. \cite{chen2024hacon} performs contrastive pretraining, clusters feature embeddings via K-means, and samples typical points from each cluster to seed the cold-start stage. Zhao et al.\citep{zhao2022ssal} similarly leverages self-supervised learning (SSL) representations in dermatology, selecting cluster centroids and their nearest neighbors to ensure representativeness. COLoSSAL \citep{liu2023colossal} provides a large-scale benchmark for cold-start AL in 3D segmentation, confirming that TypiClust \citep{hacohen2022typiclust}, which clusters SSL embeddings and picks the densest typical points, achieves robust low-budget performance. More recent studies further explore the use of foundation models for feature-space extraction in clustering-based selection \citep{yuan2024fmcluster, yuan2025rethinking}.

When an initial network is available and the objective is to select additional samples, uncertainty-based strategies are commonly used, prioritizing cases for which the model exhibits the highest uncertainty. Uncertainty-based active learning assumes that samples with ambiguous predictions are more informative for improving the model and commonly relies on criteria such as least confidence, margin sampling, or predictive entropy. In deep networks, uncertainty is often estimated using Bayesian approximations such as Monte Carlo dropout, test-time augmentations, or deep ensembles, enabling principled acquisition functions like BALD or entropy-based querying \cite{gal2017deep,wang2024comprehensive}. In medical image segmentation, uncertainty can be computed at the pixel or voxel level and aggregated to guide case-level selection \cite{nath2020diminishing,specktor2023test}, with multiple studies demonstrating reduced annotation effort and competitive performance compared to random sampling \cite{kendall2017uncertainties}.

Other approaches explicitly integrate diversity with uncertainty to balance exploration and exploitation. Rokach et al. (2025) \citep{rokach2025deepal} propose a dynamic deep AL framework for abdominal CT that computes per-image uncertainty via entropy and margin scores, then injects diversity by uniformly sampling among the \(\beta\%\) top-uncertain candidates rather than greedily taking maxima. Similarly, and closest to our approach, Yuan et al. \citep{yuan2024fmcluster,yuan2025rethinking} initialize annotation using K-means clustering on foundation-model embeddings to ensure representational diversity, while later rounds select samples based on mean pixel-wise uncertainty, achieving faster convergence and improved downstream clinical segmentation and classification performance.  Beyond medical imaging, several batch-mode AL formulations also combine uncertainty and diversity explicitly, for example, via mutual-information batch acquisition \citep{kirsch2019batchbald} or by selecting diverse, high-magnitude gradient embeddings \citep{ash2019deep}.

Despite these advances, prior works often treat diversity and uncertainty as separate objectives or combine them heuristically, without offering transparency or user control over their interaction. In contrast, we propose an intuitive and interpretable framework that enables visualization of the feature-space distribution of candidate samples and introduces a simple parameter to balance diversity and uncertainty during selection. This allows practitioners to both understand and steer the sampling process, leading to more balanced and efficient cold-start initialization.

\section{Methodology}
\begin{figure}[!t]
\includegraphics[width=\textwidth,height=0.35\textheight]{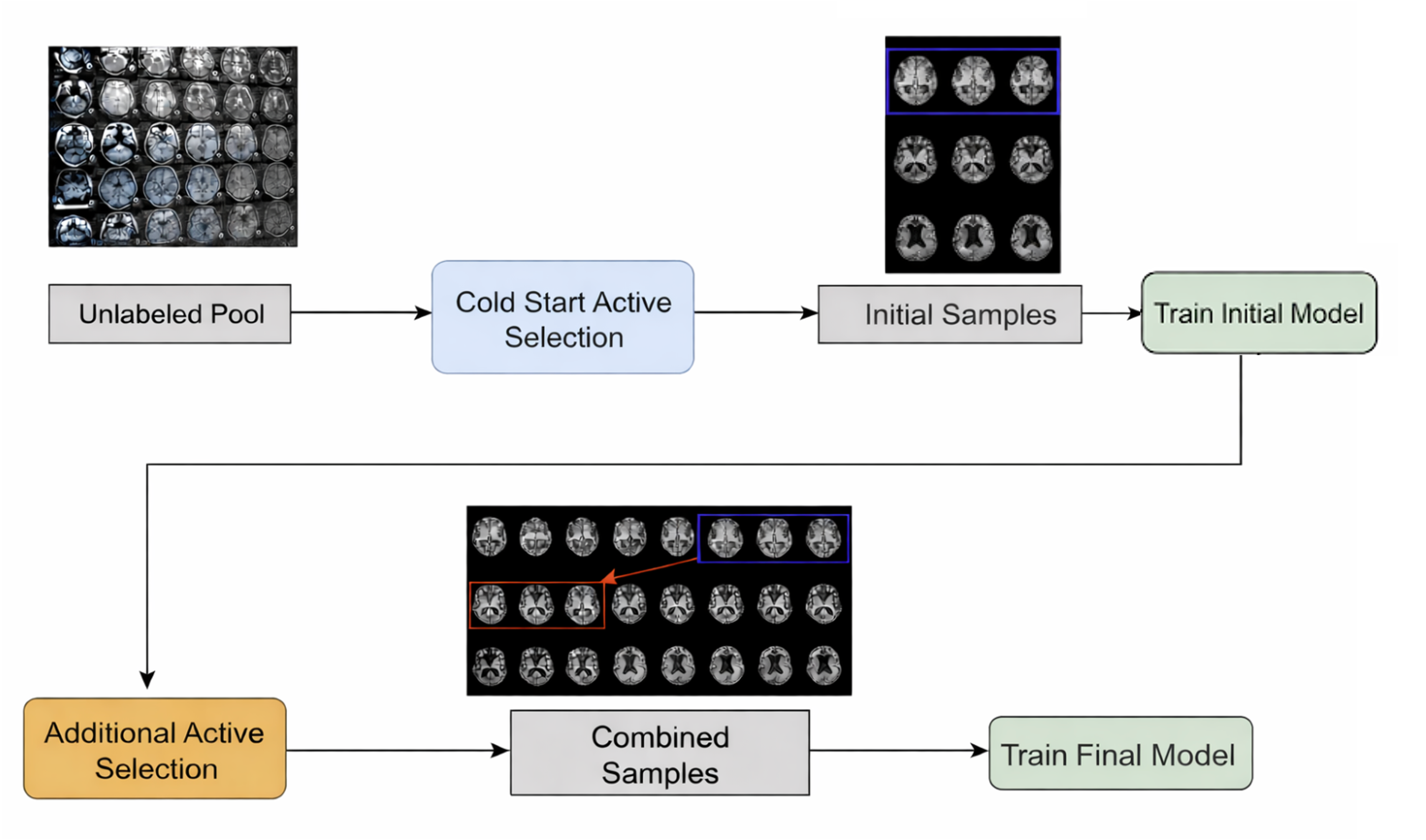}
\caption{Framework overview: cold-start followed by active selection}
\label{fig:Methodology_overview}
\end{figure}

Considering an unlabeled pool of medical images $\mathcal{U}=\{x_i\}_{i=1}^{N}$, the objective is to select a small subset $\mathcal{S}\subset\mathcal{U}$ for annotation that maximizes downstream segmentation performance under a fixed labeling budget. Our method covers both the construction of a cold start set and an active selection of additional cases.
Our methodology follows a two-stage process illustrated in Figure~\ref{fig:Methodology_overview}. We first obtain feature embeddings from a pre-trained foundation model, reduce their dimensionality, and cluster them to identify representative samples for cold start initialization. The labeled set is then expanded using the model trained on the initial set through active selection, where we combine image-level uncertainty with spatial diversity to guide further acquisitions.

\subsection{Cold-Start Initialization}
\begin{figure}[!t]
\includegraphics[width=\textwidth,height=0.47\textheight]{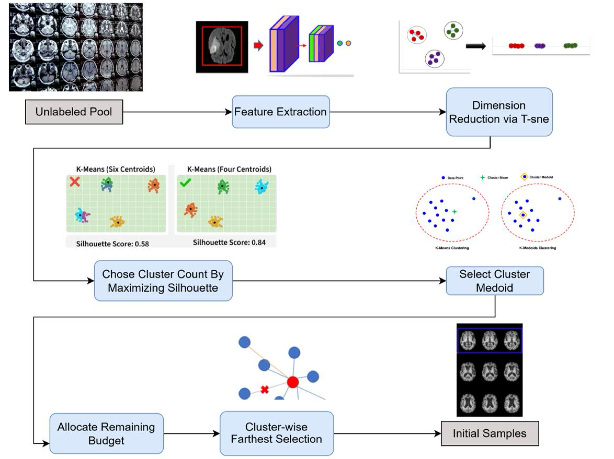}
  \caption{Our cold-start method. First, foundation-model embeddings are extracted and projected to 2D using t-SNE; the optimal number of clusters is then selected via silhouette-scored k-means, cluster medoids are chosen as initial seeds, and the remaining budget is filled by farthest-point sampling within clusters to maximize diversity.}
  \label{fig:Methodology_cold_start}
\end{figure}

Figure~\ref{fig:Methodology_cold_start} illustrates the full cold-start pipeline. First, we extract feature embeddings using a foundation model and project them into a 2D space using t-SNE for efficient distance computation. We then run k-means with several candidate values of k and select the configuration with the highest silhouette score. From the chosen clustering, we identify the medoid of each cluster as an initial representative. The remaining annotation budget is allocated proportionally to cluster size, and within each cluster we iteratively select the farthest point from the already-selected set to maximize diversity.
\subsubsection{Feature extraction.}
Each image is embedded using a strong visual encoder to obtain semantically meaningful feature representations for sample selection. We use a medical foundation model, i.e., a large-scale vision encoder adapted to medical imaging. In our implementation, a ResNet-50 encoder pretrained on the RadImageNet database is employed. RadImageNet contains 1.35 million annotated CT, MRI, and ultrasound images from 131,872 patients spanning musculoskeletal, neurologic, oncologic, gastrointestinal, endocrine, abdominal, and pulmonary pathologies~\citep{mei2022radimagenet,warvito_radimagenet_models}.

\subsubsection{Dimensionality reduction.}
In our pipeline we project encoder features to a 2D t\text{-}SNE \citep{vanDerMaaten2008tsne} map and perform both clustering and distance-based selection in this space. We chose this for pragmatic reasons: 
(i) The 2D layout is far more intuitive to inspect than the original high-dimensional features, making the selection behavior transparent; 
(ii) Computing pairwise distances and running $k$-means is computationally simpler in 2D;
(iii) t\text{-}SNE preserves local neighborhoods, which is the regime our diversity and farthest-first criteria rely on (i.e., distinguishing near/very-similar samples). We note that the same selection principles are not tied to t\text{-}SNE and can be applied in the original embedding space or with other reductions (e.g., PCA). Thus, t\text{-}SNE serves as a practical, human-interpretable proxy for local structure, while the framework itself remains agnostic to the specific dimensionality-reduction method.

\subsubsection{Choosing the number of clusters.}
We cluster the 2D embeddings with $k$-means \citep{macqueen1967kmeans}. Multiple values of $k$ are evaluated, and we select the one that maximizes the mean silhouette score \citep{rousseeuw1987silhouettes}:
\begin{equation}
\hat{k} \;=\; \arg\max_{k\in\mathcal{K}} \; S(k),
\end{equation}
where $S(k)$ is the average silhouette coefficient, which quantifies how well each sample fits within its assigned cluster by comparing its cohesion (similarity to points in the same cluster) to its separation (distance from points in neighboring clusters). Higher values indicate more compact and better-separated clusters.

\subsubsection{Seeds via medoids.}
For each cluster $C_c$ ($c=1,\dots,\hat{k}$), we select the medoid 
\begin{equation}
m_c \;=\; \arg\min_{i\in C_c} \sum_{j\in C_c} \|{z}_i-{z}_j\|_2,
\end{equation}
where ${z}_i$ is the 2D t-SNE coordinate of $x_i$. The medoids form the initial seeds.

\subsubsection{Budget-proportional augmentation.}
Given a cold-start budget $B$, the remaining slots after picking all medoids are
\begin{equation}
R \;=\; B - \hat{k}.
\end{equation}
We allocate $R$ across clusters in proportion to the size of the cluster:
\begin{equation}
r_c \;=\; \operatorname{round}\!\Big(R \cdot \frac{|C_c|}{\sum_{u=1}^{\hat{k}} |C_u|}\Big)
\qquad \text{with } \sum_{c=1}^{\hat{k}} r_c \approx R
\end{equation}
adjusting by one where necessary to ensure that the total is equal to $R$. Within each cluster, additional samples are chosen by greedy farthest-point selection in the 2D space: starting from $\{m_c\}$, repeatedly adding the point that maximizes its minimum distance to the already selected set.

\subsection{Active Selection: Entropy with Spatial Diversity}
\begin{figure}[!t]
\includegraphics[width=\textwidth,height=0.65\textheight]{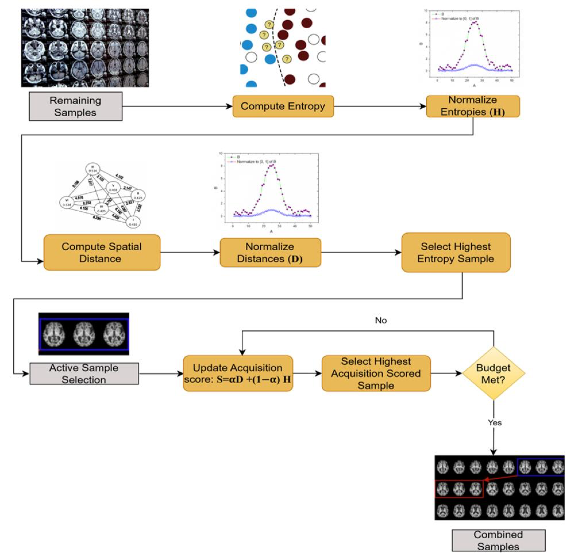}
\caption{Our combined active selection method. Entropy and diversity are computed separately and combined with a unified score.}
\label{fig:active_selection}
\end{figure}

This step follows the training of an initial network. Figure~\ref{fig:active_selection} shows the pipeline of the active selection method. First, entropy is computed for all remaining unlabeled samples and normalized. In parallel, pairwise distances between samples are evaluated in the 2D embedding space. The sample with the highest entropy is selected first, and the subsequent selections follow a mixed criterion that combines entropy with spatial diversity, favoring informative yet non-redundant samples. This iterative process continues until the acquisition budget is reached.

\subsubsection{Uncertainty (image-level entropy).}
Let $p_{i,c}(u)$ be the predicted probability at pixel $u$ for class $c$. The pixel entropy is:
\begin{equation}
H_i(u) = -\sum_c p_{i,c}(u)\log\big(p_{i,c}(u)+\epsilon\big)
\end{equation}
and the mean entropy at the image level is
\begin{equation}
\mathcal{H}(x_i) \;=\; \frac{1}{|\Omega|}\sum_{u\in\Omega} H_i(u),
\end{equation}
where $\Omega$ denotes the set of all pixel locations in the image. Thus, $\mathcal{H}(x_i)$ measures the average predictive uncertainty across the entire image.
We min-max normalize entropies between 0 and 1 over the unlabeled pool:
\begin{equation}
\tilde{\mathcal{H}}(x_i) \;=\; 
\frac{\mathcal{H}(x_i)-\min_j \mathcal{H}(x_j)}
{\max_j \mathcal{H}(x_j)-\min_j \mathcal{H}(x_j)+\varepsilon}.
\end{equation}

\subsubsection{Spatial diversity (in embedding space).}
Let $\mathcal{S}$ be the set of images already selected, and let $\mathcal{C}=\mathcal{U}\setminus\mathcal{S}$ denote the current pool of candidate images. For any candidate $i \in \mathcal{C}$, we quantify its spatial diversity by the distance to its nearest selected neighbor in the 2D embedding:
\begin{equation}
D(i \mid \mathcal{S}) \;=\; \min_{j\in \mathcal{S}} \;\|{z}_i - {z}_j\|_2.
\end{equation}
To keep diversity scores on a comparable scale, we normalize by the maximum pairwise distance observed within the candidate pool.
\begin{equation}
D_{\max} \;=\; \max_{p,q\in \mathcal{C}} \;\|{z}_p - {z}_q\|_2
\qquad
\tilde{D}(i) \;=\; \frac{D(i \mid \mathcal{S})}{D_{\max}+\varepsilon}.
\end{equation}

\subsubsection{Acquisition score}
We combine the two signals with a single trade-off parameter $\alpha\in[0,1]$:
\begin{equation}
S(i) \;=\; \alpha\,\tilde{D}(i) \;+\; (1-\alpha)\,\tilde{\mathcal{H}}(x_i).
\label{diversity_entropy_eq}
\end{equation}
At each step, we add the unlabeled image with the highest $S(i)$, updating $\mathcal{S}$, recomputing $\tilde{D}(i)$ and repeating until the desired number of images is selected. The larger $\alpha$ emphasizes coverage (discouraging selections that are close to the embedding), while the smaller $\alpha$ prioritizes uncertainty.

\section{Experimental results}
The experimental section is organized as follows. We first describe the datasets and data splits, followed by the preprocessing steps, model training, and evaluation metrics. Finally, we present a detailed description of the results.

\subsection{Datasets}
\subsubsection{Datasets description}
The method was evaluated on three distinct datasets covering different segmentation tasks: brain structures, lung, and heart. It was applied to 2D data, spanning two fundamentally different imaging modalities: X-ray and MRI. We next describe each dataset.

\textit{SynthStrip 2D dataset.} The 2D synthStrip data set \citep{hoopes2022synthstrip} was used with brain substructures annotations. This dataset contains substructures annotations for T1-weighted images: 38 FreeSurfer maintenance (FSM) images, 43 arterial-spin-labeling (ASL) images, and 50 images from the IXI dataset. This resulted in a total set of 131 images, each with multi-class anatomical labels of brain substructures (14–18 classes per slice, depending on structures present).

\textit{Montgomery Chest X-ray (CXR).}
The Montgomery County CXR set provides posterior–anterior radiographs with expert lung masks, consisting of three classes: right lung, left lung, and background \citep{jaeger2013automatic, candemir2013lung}. The dataset contains 138 x-rays, of which 80 x-rays are normal and 58 x-rays are abnormal with manifestations of tuberculosis.

\textit{NIH ChestX-ray8 with CheXmask.}
The CheXmask dataset was constructed by aggregating chest X-ray images from multiple public databases and generating standardized anatomical segmentation masks using an automated deep learning pipeline with quality control \citep{gaggion2024chexmask}. We sample 800 random frontal CXRs from ChestX-ray8 \citep{Wang_2017} and pair each with its CheXmask segmentation (left lung, right lung, heart). Using CheXmask’s per-image quality estimates via mean RCA Dice, we rank the 800 images and construct a 300-images subset from the top of this ranking. When patient identifiers are available, we keep at most one image per patient to reduce potential leakage from near-duplicate studies.

\begin{table}[t]
\centering
\setlength{\tabcolsep}{3pt} % default is ~6pt; make smaller
\renewcommand{\arraystretch}{1.2}
\begin{tabular}{|c|c|c|c|}
\hline
Data Set Name & Segmentation Task & \# Images & Test Size \\
\hline
SynthStrip T1 (2D) & Brain Structures & 131 & 26 \\
Montgomery Chest X-ray & Lungs & 138 & 28 \\
ChestX-ray8 with CheXmask-300 & Lungs and Heart & 300 & 60 \\
\hline
\end{tabular}
\caption{Sythstrip, Montgomery, and CheXmask datasets Summary.}
\end{table}

\subsubsection{Data splits}
As none of the data sets provides a fixed split, we randomly sampled 20\% of the images as a holdout test set and removed it from all selection and training procedures. The remaining 80\% constitutes the pool for cold start and subsequent active learning.
For each dataset, we perform $11$ independent runs with seeds $s_r = 42 + r$, $r \in \{1,\dots,11\}$. Each $s_r$ fixes the $20\%$ test split and is reused for all stochastic components (\(k\)-mean, silhouette evaluation, farthest point sampling, and NumPy /PyTorch initializers). In every round, models are retrained from scratch and deterministically initialized with $s_r$; thus, within a run, all models share identical initial weights and differ only in the labeled set.

\subsection{Data preprocessing and model training}
For comparability and computational efficiency, we resize every image and its mask to \(256\times256\) pixels. Apart from resizing, we apply \emph{z}-score intensity normalization to each image:
\begin{equation}
I' \;=\; \frac{I - \mu}{\sigma + \varepsilon},
\end{equation}
where \(\mu\) and \(\sigma\) are the per-image mean and standard deviation (and \(\varepsilon\) is a small constant for stability). 

All models are implemented with the MONAI framework\citep{cardoso2022monai}. 
In our experiments we used Attention U-Net \citep{oktay2018attentionunet} trained with the combined Dice and Cross entropy loss \citep{MA2021102035}, Adam \citep{kingma2015adam} optimizer, a batch size of 2 and a learning rate of $\eta=10^{-3}$. 
For chest X-ray datasets (Montgomery and CheXmask), we train for 120 epochs with dropout \(p=0.15\). 
Because of the complexity of the brain anatomy task, for the Synthstrip dataset we train for 150 epochs with dropout \(p=0.10\). 
Across active-learning rounds we retrain the model from scratch on the currently labeled set (i.e., no warm-start fine-tuning across rounds). The experiments were run in Google Colab with an NVIDIA A100 GPU.

\subsection{Evaluation}
Performance is evaluated on the fixed 20\% test set. The primary metric is the Dice coefficient. We report the per-class Dice score averaged across structures for the chest X-ray datasets (Montgomery and CheXmask). Owing to the presence of classes that are absent or highly underrepresented in the training folds of the SynthStrip dataset, for this dataset we use image-level Dice instead of per-class Dice. We also report the 95th percentile Hausdorff distance (HD95), which computes the Hausdorff distance at the 95th percentile to reduce sensitivity to outliers \citep{muller2022metrics}.

\subsection{Results}
We compare our pipeline against strong baselines for both the cold start and the active selection settings. All methods are matched on total annotation budget and use the same train/test splits, seeds, and training protocol. The number of training examples is adapted for each dataset to account for differences in task difficulty. Specifically, we use between 13 and 39 training examples for the SynthStrip dataset, 8 to 26 training examples for the Montgomery dataset, and 9 to 18 training examples for the CheXmask dataset.

\textit{Cold start.} Let $B$ be the number of training samples. For the cold-start setting we use the following algorithm variants:
\begin{itemize}
  \item Random seeding. Uniformly sample $B$ images from the 80\% pool; no clustering.
  \item KMeans in the feature space. Run $k$-means with $k=B$ on encoder features $f_\theta(x)$ and pick the medoid, following the approach of Yuan et al.  \citep{yuan2025rethinking}. This approach is also related to TypiClust \citep{hacohen2022typiclust}, but replaces self-supervised representations with features extracted from a foundation model. 
  \item Our (cold-start). RadImageNet–ResNet50 features, t-SNE embedding, $\hat{k}$ via silhouette, medoids andproportional farthest-point augmentation.
\end{itemize}

\textit{Active selection policies on top of the cold-start.}
Let the total annotation budget be \(B\), of which \(A\) images are acquired during active learning; the cold-start set therefore has size \(B_0 = B - A\).
At each selection step, we train a segmentation model \(m\) from scratch on \(B_0\).
Given \(m\), the acquisition score for an unlabeled image \(x_i\) is calculated using equation \eqref{diversity_entropy_eq} to select additional \(A\) images. We test different entropy-diversity combinations for the selection:
\begin{itemize}
  \item Pure Entropy. Only uncertainty term calculated by top image-level entropy. For the entropy-only setting, we set \(\alpha=0\).
  \item Our combination of uncertainty and diversity with a single tunable trade-off; we use \(\alpha=0.3\) by default.
  \item Pure diversity. Only spatial diversity term using \(\alpha=1\). Uncertainty is used only to select the initial single example, followed by diversity based selection similar to Geifman and El-Yaniv \citep{geifman2017ffactive}.
\end{itemize}

\subsubsection{Results on the SynthStrip 2D dataset}

\begin{figure}[t]
  \centering
  \includegraphics[width=\textwidth,height=0.9\textheight,keepaspectratio]{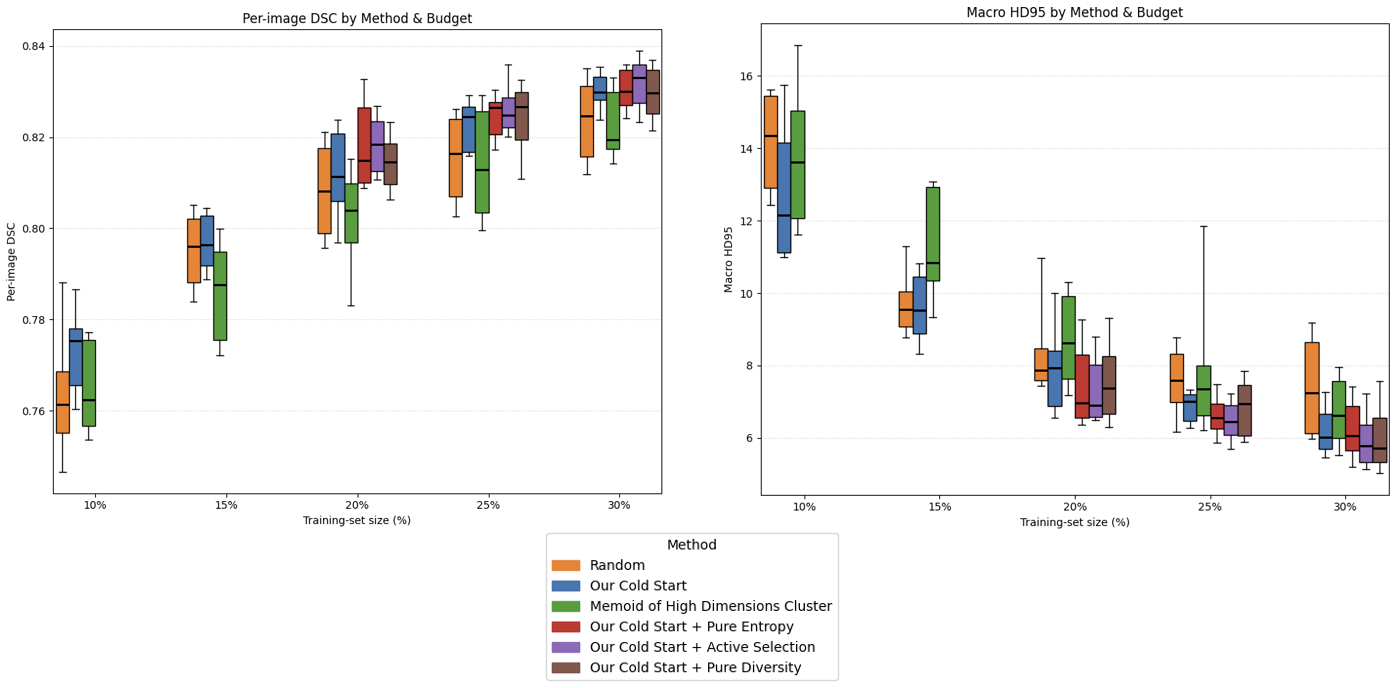}
  \caption{SynthStrip 2D Result Summary}
  \label{fig:synthstrip_results}
\end{figure}

The Synthstrip dataset comprises 131 brain MRI scans, with each split using 26 for testing, leaving 105 potential training cases. Comprehensive results across all annotation budgets are presented in Figure \ref{fig:synthstrip_results} and Table~\ref{tab:Synthstrip_results}.

\begin{table}[H]
\caption{Results on the SynthStrip dataset for different annotation budgets. The comparison includes random selection, cold-start clustering (ours), medoid-based selection, and active learning strategies using entropy ($\alpha=0$), a combination of entropy and distance ($\alpha=0.3$), and distance only ($\alpha=1$). Active learning methods are evaluated only for the 20\%, 25\%, and 30\% budgets.}

\setlength{\tabcolsep}{2pt}
\renewcommand{\arraystretch}{1.2}
\setlength{\extrarowheight}{1.5pt}
\large % try \normalsize if it still fits
\resizebox{\textwidth}{!}{%
\begin{tabular}{|cllcccc|}
\hline
Budget (\#)                & Setting                     & Method                       & \begin{tabular}[c]{@{}c@{}}Dice ↑\\ (mean±std)\end{tabular} & \begin{tabular}[c]{@{}c@{}}Dice ↑\\ (median)\end{tabular} & \begin{tabular}[c]{@{}c@{}}HD95 ↓\\ (mean±std)\end{tabular} & \begin{tabular}[c]{@{}c@{}}HD95 ↓\\ (median)\end{tabular} \\ \hline
\multirow{3}{*}{10\% (13)} & \multirow{3}{*}{Cold Start} & Random-only                  & 0.7637±0.01                                                 & 0.761                                                     & 14.26±1.95                                                  & 14.35                                                     \\
                           &                             & Cold-start clustering (ours) & \textbf{0.7732±0.01}                                        & \textbf{0.775}                                            & \textbf{12.8±1.93}                                          & \textbf{12.15}                                            \\
                           &                             & K-means to budget  & 0.7651±0.01                                                 & 0.762                                                     & 13.9±2.19                                                   & 13.61                                                     \\ \hline
\multirow{3}{*}{15\% (19)} & \multirow{3}{*}{Cold start} & Random-only                  & 0.7954±0.009                                                & 0.796                                                     & \textbf{9.64±1.3}                                           & 9.54                                                      \\
                           &                             & Cold-start clustering (ours) & \textbf{0.7963±0.008}                                       & 0.796                                                     & 9.66±1.31                                                   & \textbf{9.51}                                             \\
                           &                             & K-means to budget  & 0.7868±0.01                                                 & 0.788                                                     & 11.28±1.66                                                  & 10.83                                                     \\ \hline
\multirow{6}{*}{20\% (26)} & \multirow{3}{*}{Cold start} & Random-only                  & 0.8082±0.01                                                 & 0.808                                                     & 8.37±1.32                                                   & 7.87                                                      \\
                           &                             & Cold-start clustering (ours) & 0.8119±0.01                                                 & 0.811                                                     & 7.93±1.21                                                   & 7.94                                                      \\
                           &                             & K-means to budget  & 0.8021±0.01                                                 & 0.804                                                     & 8.66±1.52                                                   & 8.61                                                      \\ \cline{2-7} 
                           & \multirow{3}{*}{AL13+13}    & Cluster+Entropy $\alpha±=0.0$     & 0.8185±0.009                                                & 0.815                                                     & 7.48±1.14                                                   & 6.95                                                      \\
                           &                             & Cluster+Combined $\alpha±=0.3$     & \textbf{0.8188±0.006}                                       & \textbf{0.819}                                            & \textbf{7.29±1.05}                                          & \textbf{6.9}                                              \\
                           &                             & Cluster+Distance $\alpha±=1.0$     & 0.8136±0.008                                                & 0.815                                                     & 7.59±1.1                                                    & 7.37                                                      \\ \hline
\multirow{6}{*}{25\% (33)} & \multirow{3}{*}{Cold start} & Random-only                  & 0.8157±0.009                                                & 0.816                                                     & 7.58±0.98                                                   & 7.54                                                      \\
                           &                             & Cold-start clustering (ours) & 0.8224±0.006                                                & 0.824                                                     & 6.83±0.56                                                   & 7                                                         \\
                           &                             & K-means to budget  & 0.8146±0.01                                                 & 0.813                                                     & 8.02±2.18                                                   & 7.36                                                      \\ \cline{2-7} 
                           & \multirow{3}{*}{AL 20+13}   & Cluster+Entropy $\alpha±=0.0$      & 0.8245±0.006                                                & 0.826                                                     & 6.68±1.04                                                   & 6.56                                                      \\
                           &                             & Cluster+Combined $\alpha±=0.3$     & \textbf{0.8259±0.006}                                       & 0.825                                                     & \textbf{6.5±0.82}                                           & \textbf{6.44}                                             \\
                           &                             & Cluster+Distance $\alpha±=1.0$     & 0.8243±0.003                                                & \textbf{0.827}                                            & 6.77±1.08                                                   & 6.65                                                      \\ \hline
\multirow{6}{*}{30\% (39)} & \multirow{3}{*}{Cold start} & Random-only                  & 0.8239±0.008                                                & 0.825                                                     & 7.32±1.39                                                   & 7.25                                                      \\
                           &                             & Cold-start clustering (ours) & 0.8296±0.006                                                & 0.830                                                     & 6.21±0.67                                                   & 6.01                                                      \\
                           &                             & K-means to budget  & 0.8231±0.008                                                & 0.820                                                     & 6.78±1.14                                                   & 6.63                                                      \\ \cline{2-7} 
                           & \multirow{3}{*}{AL 26+13}   & Cluster+Entropy $\alpha±=0.0$      & 0.8306±0.006                                                & 0.830                                                     & 6.23±0.88                                                   & 6.06                                                      \\
                           &                             & Cluster+Combined $\alpha±=0.3$     & \textbf{0.8325±0.005}                                       & \textbf{0.833}                                            & \textbf{5.98±0.83}                                          & 5.79                                                      \\
                           &                             & Cluster+Distance $\alpha±=1.0$     & 0.8295±0.008                                                & 0.830                                                     & 6.22±1.35                                                   & \textbf{5.71}                                             \\ \hline
\end{tabular}}
\label{tab:Synthstrip_results}
\end{table}

Across all labeling budgets, our cold-start clustering consistently matches or outperforms random initialization, whereas the naive k-means-to-budget variant underperforms even relative to random selection. Specifically, our method achieves an average Dice score of 0.807, compared to 0.801 for random selection and 0.798 for the k-means-to-budget approach. In terms of Hausdorff distance, it attains 8.686 mm, outperforming random selection (9.434 mm) and k-means-to-budget (9.728 mm). We hypothesize that this behavior arises from directly tying the number of clusters, K, to a relatively large budget, which can lead to over-aggregation of dominant modes and the selection of outliers as cluster representatives.

In the active learning stage (budgets 20\%, 25\% and 30\%), all active selection strategies outperform baselines. Among them, our acquisition combined (entropy and distance) is slightly but consistently the strongest. This approach achieves an overall Dice score of 0.826, compared to 0.816 for random selection, 0.821 for our cold-start initialization, 0.813 for the k-means-to-budget method, 0.825 for pure entropy, and 0.822 for pure distance. In terms of Hausdorff distance, our combined method reaches 6.59 mm, outperforming random selection (7.76 mm), cold start (6.99 mm), k-means-to-budget (7.82 mm), pure entropy (6.80 mm), and pure distance (6.86 mm).

An important observation concerns the differences in standard deviation (STD), which reflect the variability of the results across runs. Overall, random selection exhibits higher variability, with a Dice STD of 0.009 and a Hausdorff STD of 1.388, compared to a Dice STD of 0.008 and a Hausdorff STD of 1.136 achieved by our cold-start method. In the active selection setting, random selection again shows greater variability, with a Dice STD of 0.009 and a Hausdorff STD of 1.23, whereas our combined uncertainty–diversity strategy achieves lower variability, with a Dice STD of 0.006 and a Hausdorff STD of 0.9.

\begin{figure}[t]
  \centering
  \includegraphics[width=\textwidth,height=0.9\textheight,keepaspectratio]{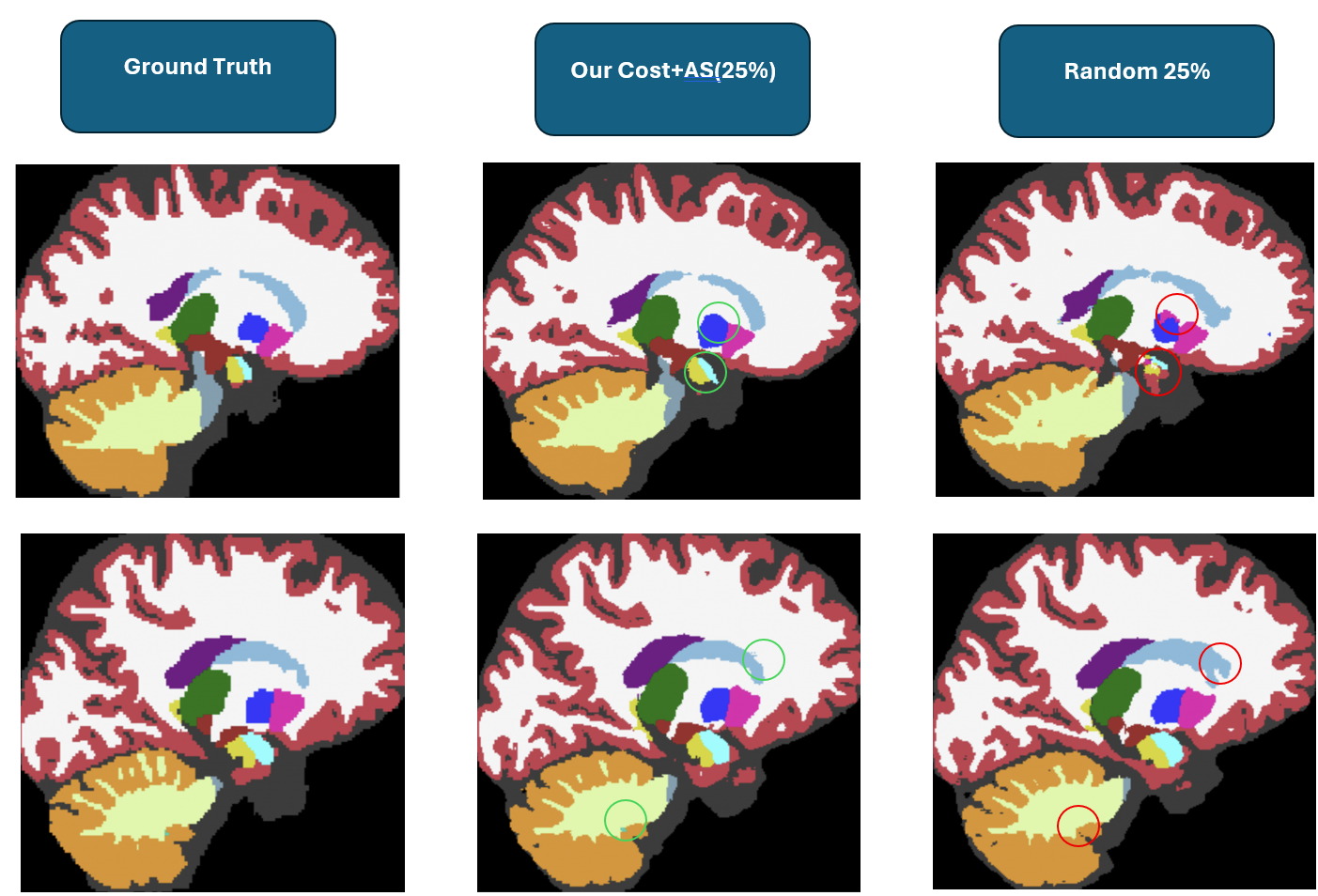}
  \caption{Illustrative Examples for results comparison between random selection and our active selection method that combined entropy with diversity. Large differences are emphasized with green and red circles.}
  \label{fig:illustrative_examples}
\end{figure}

Figure~\ref{fig:illustrative_examples} presents illustrative qualitative examples comparing the combined active selection method with random selection, with circles highlighting the differences. Our method yields more accurate boundary delineation and demonstrates improved handling of small structures.

\begin{figure}[t]
  \centering
  \includegraphics[width=\textwidth,height=0.9\textheight,keepaspectratio]{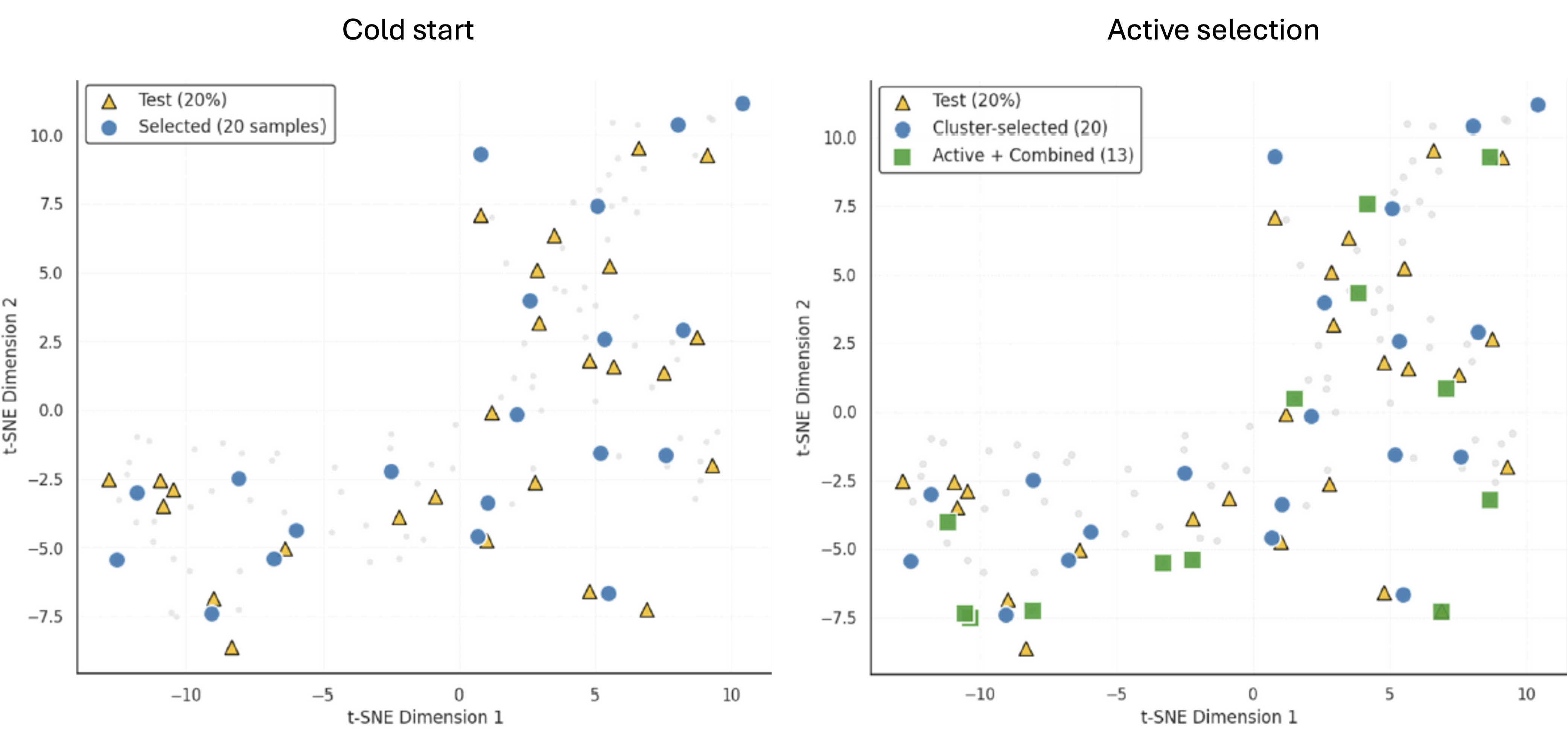}\\
  \caption{Example of T-SNE visualization on the selected cases. Cold start selection on the left; active selection when the initial network is available on the right. Blue points denote the initially selected samples, green squares indicate actively selected cases using the proposed combined method, orange triangles represent test samples, and small gray dots correspond to the unlabeled pool.}
  \label{fig:t-sne}
\end{figure}

Figure \ref{fig:t-sne} illustrates the t-SNE embedding of the selected cases. The initially selected samples are shown as blue points, the actively selected cases obtained using our combined method are indicated by green squares, and the test samples are represented by orange triangles. The unlabeled pool from which the training samples were selected is shown as small gray dots. Overall, the selected cases provide good coverage of the test-set feature space.

\subsubsection{Results on the Montgomery dataset}

\begin{figure}[t]
  \centering
  \includegraphics[width=\textwidth,height=0.9\textheight,keepaspectratio]{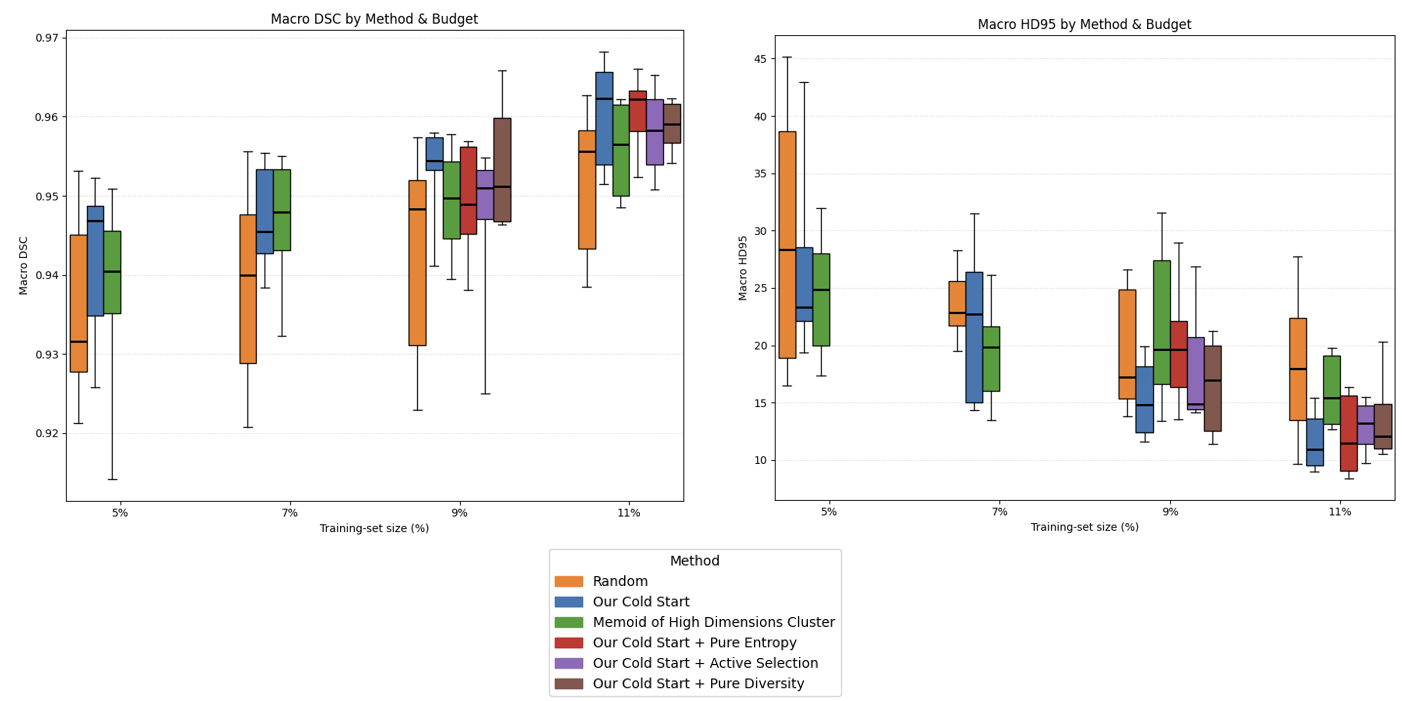}
  \caption{Montgomery Result Summary}
  \label{fig:Montgometry_results}
\end{figure}

This chest x-ray dataset contains 138 images, 28 reserved for testing, leaving 110 potential training cases. The lungs segmentation task is relatively simple, therefore we use small labeling budgets starting from 8 labeled training examples. The full quantitative results for all annotation budgets are reported in Figure ~\ref{fig:Montgometry_results} and Table~\ref{tab:Montgometry_results}.

 Across budgets, our cold start clustering consistently matches or exceeds random initialization and is more stable than the "K-means to budget" variant. It results in an average Dice score of 0.95 compared to 0.928 for random selection and 0.945 for the "K-means to budget" method, as well as a reduced average Hausdorff distance of 20.44 mm compared to 25.2 mm of random selection and 21.33 mm of the "K-means to budget" method. Among the informed strategies, no single variant dominates across all budgets, but they almost always avoid outliers and deliver consistently higher segmentation quality compared to random selection.

\begin{table}[H]
\caption{Results on the Montgomery dataset for different annotation budgets. The comparison includes random selection, cold-start clustering (ours), medoid-based selection, and active learning strategies based on entropy ($\alpha=0$), a combination of entropy and distance ($\alpha=0.3$), and distance only ($\alpha=1$). For 8\% and 10\% budgets, only cold-start methods are compared.}
%\begingroup
\setlength{\tabcolsep}{2pt}
\renewcommand{\arraystretch}{1.20}
\setlength{\extrarowheight}{1.5pt}
\large 
\resizebox{\textwidth}{!}{%
\begin{tabular}{|cllcccc|}
\hline
Budget (\#)                & Setting                     & Method                       & \begin{tabular}[c]{@{}c@{}}Dice↑ \\ (mean)\end{tabular} & \begin{tabular}[c]{@{}c@{}}Dice↑ \\ (median)\end{tabular} & \begin{tabular}[c]{@{}c@{}}HD95↓ \\ (mean)\end{tabular} & \begin{tabular}[c]{@{}c@{}}HD95↓\\ (median)\end{tabular} \\ \hline
\multirow{3}{*}{5\% (8)}   & \multirow{3}{*}{Cold start} & Random-only                  & 0.9297±0.02                                             & 0.9316                                                    & 30.77±14.02                                             & 28.37                                                    \\
                           &                             & Cold-start clustering (ours) & \textbf{0.9413±0.01}                                    & \textbf{0.9468}                                           & 26.78±9.32                                              & \textbf{23.33}                                           \\
                           &                             & K-means to budget & 0.9371±0.01& 0.9405                                                    & \textbf{25.29±8.29}                                     & 24.85                                                    \\ \hline
\multirow{3}{*}{7\% (10)}  & \multirow{3}{*}{Cold start} & Random-only                  & 0.9399±0.01                                             & 0.9384                                                    & 23.73±4                                                 & 22.83                                                    \\
                           &                             & Cold-start clustering (ours) & 0.9454±0.02                                             & 0.9408                                                    & 24.81±14.08                                             & 22.74                                                    \\
                           &                             & K-means to budget  & \textbf{0.9459±0.009}                                   & \textbf{0.9479}                                           & \textbf{19.57±5.38}                                     & \textbf{19.85}                                           \\ \hline
\multirow{6}{*}{9\% (12)}  & \multirow{3}{*}{Cold start} & Random-only                  & 0.8929±0.16                                             & 0.9483                                                    & 28.34±32.03                                             & 17.25                                                    \\
                           &                             & Cold-start clustering (ours) & \textbf{0.9524±0.009}                                   & \textbf{0.9544}                                           & 18.2±10.7                                               & \textbf{14.79}                                           \\
                           &                             & K-means to budget  & 0.9412±0.02                                             & 0.9497                                                    & 24.12±12.67                                             & 19.64                                                    \\ \cline{2-7} 
                           & \multirow{3}{*}{AL 8+4}     & Cluster+Entropy $\alpha=0.0$        & 0.9224±0.08                                             & 0.9489                                                    & 27.57±27.39                                             & 19.63                                                    \\
                           &                             & Cluster+Combined $\alpha=0.3$       & 0.9343±0.04                                             & 0.951                                                     & 21.08±13.82                                             & 14.89                                                    \\
                           &                             & Cluster+Distance $\alpha=1.0$       & 0.95±0.01                                               & 0.9512                                                    & \textbf{17.58±7.42}                                     & 16.98                                                    \\ \hline
\multirow{6}{*}{11\% (16)} & \multirow{3}{*}{Cold start} & Random-only                  & 0.9505±0.01                                             & 0.9557                                                    & 17.95±6.83                                              & 18.26                                                    \\
                           &                             & Cold-start clustering (ours) & 0.9595±0.007                                            & \textbf{0.9623}                                           & \textbf{11.96±3.4}                                      & \textbf{10.93}                                           \\
                           &                             & K-means to budget  & 0.9558±0.007                                            & 0.9565                                                    & 16.34±3.27                                              & 15.43                                                    \\ \cline{2-7} 
                           & \multirow{3}{*}{AL 12+4}    & Cluster+Entropy $\alpha=0.0$        & \textbf{0.9604±0.005}                                   & 0.9622                                                    & 12.34±3.75                                              & 11.47                                                    \\
                           &                             & Cluster+Combined $\alpha=0.3$       & 0.9581±0.007                                            & 0.9583                                                    & 12.87±2.11                                              & 13.2                                                     \\
                           &                             & Cluster+Distance $\alpha=1.0$       & 0.9583±0.007                                            & 0.9591                                                    & 13.88±4.65                                              & 12.07                                                    \\ \hline
\end{tabular}}
\label{tab:Montgometry_results}
\end{table}

In the active learning stage with the 9\% and the 11\% budgets, our cold start outperforms all other methods, including active selection with an additional uncertainty term, reaching a Dice score of 0.956 compared to 0.922 for random selection, 0.9485 for the "K-means to budget" method, 0.941 for the entropy-based active selection method, 0.946 for the combined active selection using clustering and uncertainty, and 0.954 for the  active selection based on the pure distance method. In terms of Hausdorff distance, it reached the smallest distance of 15.08 mm compared to a distance of  23.14 mm for random selection, 20.23 mm for the K-means to budget method, 19.96 mm for the active selection with pure entropy, 16.97 mm for the combined active selection of entropy and distance, and 15.73 mm for active selection with pure distance.

With respect to result variability, random selection exhibits substantially higher variability overall, achieving a Dice standard deviation (STD) of 0.05 and a Hausdorff STD of 14.22, compared to a Dice STD of 0.012 and a Hausdorff STD of 9.375 obtained with our cold-start method. In the active selection setting, random selection again demonstrates higher variability, with a Dice STD of 0.085 and a Hausdorff STD of 19.43, whereas our combined uncertainty–diversity strategy yields lower variability, with a Dice STD of 0.023 and a Hausdorff STD of 7.965.

\subsubsection{Results on the CheXmask-300 dataset}
\begin{figure}[t]
  \centering
  \includegraphics[width=\textwidth,height=0.9\textheight,keepaspectratio]{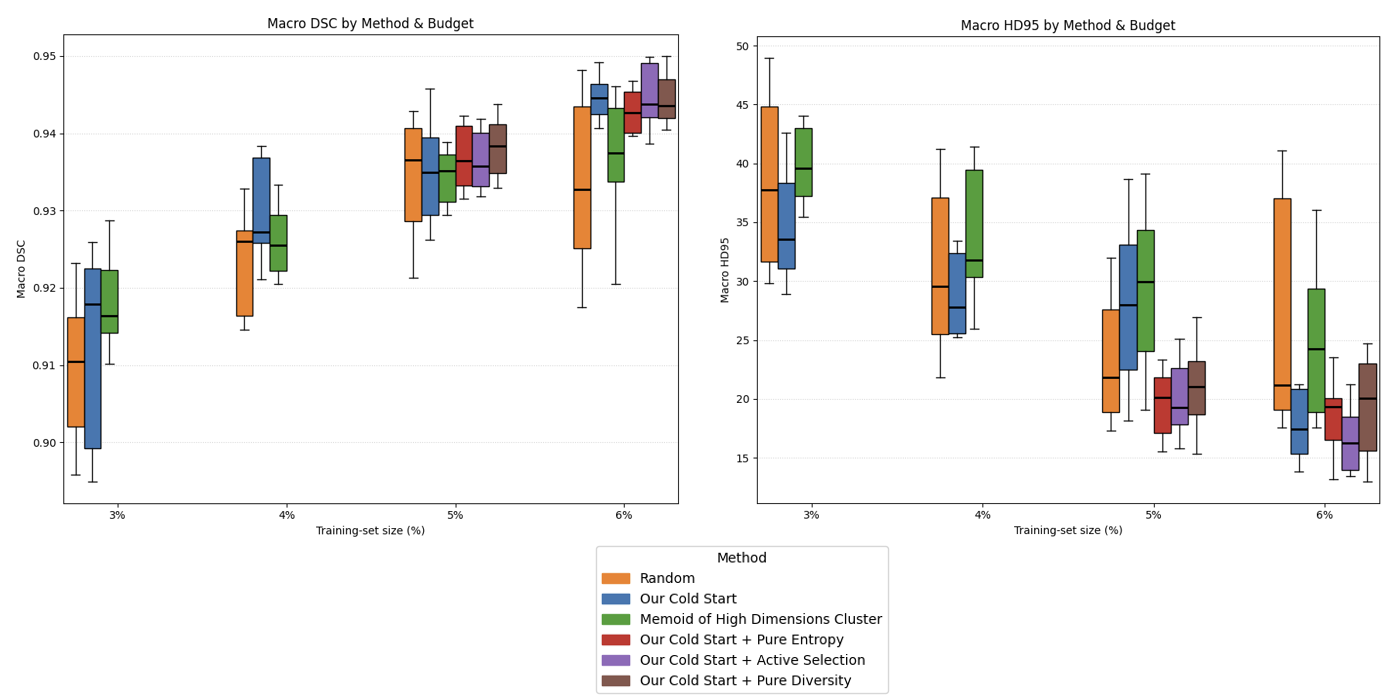}
  \caption{CheXmask-300 Result Summary}
  \label{fig:CheXmask}
\end{figure}

The CheXmask dataset comprises a larger pool of samples than the other datasets and includes an additional heart structure. It consists of 300 images, of which 60 are reserved for testing, leaving 240 candidate cases for training. As the annotations are generated by the CheXmask network, a certain level of label noise is expected. Complete quantitative results across all annotation budgets are presented in Figure~\ref{fig:CheXmask} and Table~\ref{tab:CheXmask_results}. Overall, the embedding-based initialization and the informed acquisitions reduce outliers and improve consistency on this larger pool.

Across annotation budgets, our cold-start clustering demonstrates greater stability than the baseline methods and frequently improves over random selection. It achieves an average Dice score of 0.929, compared to 0.918 for random selection and 0.923 for the k-means-to-budget method, and an average Hausdorff distance of 27.66 mm, outperforming random selection (32.41 mm) and k-means-to-budget (34.24 mm).

\begin{table}[H]
\caption{Results on the CheXmask—300 dataset for different annotation budgets. The comparison includes random selection, cold-start clustering (ours), medoid-based selection, and active learning strategies based on entropy ($\alpha=0$), a combination of entropy and distance ($\alpha=0.3$), and distance only ($\alpha=1$). For 8\% and 10\% budgets, only cold-start methods are compared, while active learning methods are additionally evaluated for 12\% and 16\% budgets.}
\begingroup
\setlength{\tabcolsep}{2pt}
\renewcommand{\arraystretch}{1.2}
\setlength{\extrarowheight}{1.5pt}
\large
\resizebox{\textwidth}{!}{
\begin{tabular}{|lllcccc|}
\hline
Budget (\#)                & Setting                     & Method                       & \begin{tabular}[c]{@{}c@{}}Dice↑ \\ (mean)\end{tabular} & \begin{tabular}[c]{@{}c@{}}Dice↑ \\ (median)\end{tabular} & \begin{tabular}[c]{@{}c@{}}HD95↓ \\ (mean)\end{tabular} & \begin{tabular}[c]{@{}c@{}}HD95↓\\ (median)\end{tabular} \\ \hline
\multirow{3}{*}{3\%  (9)} & \multirow{3}{*}{Cold start} & Random-only                  & 0.9094±0.009 & 0.9104       & 38.73±7.59  & 37.78        \\
                          &                             & Cold-start clustering (ours) & 0.9124±0.01  & \textbf{0.9179}       & \textbf{34.95±4.92}  & \textbf{33.55}        \\
                          &                             & K-means to budget  & \textbf{0.9185±0.006}& 0.9164       & 40.58±5.1   & 39.61        \\ \hline
\multirow{3}{*}{4\% (12)} & \multirow{3}{*}{Cold start} & Random-only                  & 0.9229±0.008 & 0.926        & 30.72±7.73  & 29.53        \\
                          &                             & Cold-start clustering (ours) & \textbf{0.9296±0.007} & \textbf{0.9272 }      & 2\textbf{8.78±4.44}  & \textbf{27.78 }       \\
                          &                             & K-means to budget  & 0.9261±0.005 & 0.9255       & 34.07±6.35  & 31.81        \\ \hline
\multirow{6}{*}{5\% (15)} & \multirow{3}{*}{Cold start} & Random-only                  & 0.933±0.01   & 0.9366       & 25.39±10.56 & 21.84        \\
                          &                             & Cold-start clustering (ours) & 0.9311±0.01  & 0.935        & 28.64±9.49  & 27.95        \\
                          &                             & K-means to budget)  & 0.933±0.006  & 0.9352       & 29.41±7.26  & 29.79        \\ \cline{2-7} 
                          & \multirow{3}{*}{AL 10+5}    & Cluster+Entropy $\alpha=0.0$  & 0.9366±0.006 & 0.9364       & \textbf{19.42±3.29}  & 20.11        \\
                          &                             & Cluster+Combined $\alpha=0.3$ & 0.9361±0.005 & 0.9358       & 19.99±3.81  & \textbf{19.25}        \\
                          &                             & Cluster+Distance $\alpha=1.0$ & \textbf{0.9384±0.004} & \textbf{0.9384 }      & 20.98±3.87  & 21.02        \\ \hline
\multirow{6}{*}{6\% (18)} & \multirow{3}{*}{Cold start} & Random-only                  & 0.9056±0.09  & 0.9327       & 34.81±32    & 21.19        \\
                          &                             & Cold-start clustering (ours) & 0.9437±0.006 & \textbf{0.9446}       & \textbf{18.26±3.46}  & 17.42        \\
                          &                             & K-means to budget  & 0.9137±0.07  & 0.9374       & 32.91±30.57 & 24.25        \\ \cline{2-7} 
                          & \multirow{3}{*}{AL 13+5}    & Cluster+Entropy $\alpha=0.0$  & 0.9424±0.005 & 0.9427       & 19.33±5.32  & 19.35        \\
                          &                             & Cluster+Combined $\alpha=0.3$ & 0.9413±0.01  & 0.9437       & 18.33±7.23  & \textbf{16.26}        \\
                          &                             & Cluster+Distance $\alpha=1.0$ & \textbf{0.9445±0.004} & 0.9436       & 19.5±4.64   & 20.04        \\ \hline
\end{tabular}}
\endgroup
\label{tab:CheXmask_results}
\end{table}

In the active learning stage (budgets of 5\% and 6\%), all active learning strategies outperform the cold-start baselines. In terms of Dice score, the pure distance–based acquisition slightly outperforms the other active learning methods, reaching 0.941, compared to 0.919 for random selection, 0.937 for our cold-start strategy, 0.922 for the k-means-to-budget strategy, 0.940 for entropy-only acquisition, and 0.939 for the combined entropy–distance strategy. With respect to Hausdorff distance, the combined acquisition strategy yields the best performance, achieving a distance of 19.16 mm, compared to 30.10 mm for random selection, 23.45 mm for our cold-start method, 31.16 mm for the k-means-to-budget method, 19.37 mm for entropy-only acquisition, and 20.24 mm for the pure distance strategy.

With respect to result variability, random selection exhibits substantially higher variability overall, with a Dice standard deviation (STD) of 0.029 and a Hausdorff STD of 14.47, compared to the markedly lower variability achieved by our cold-start method, which attains a Dice STD of 0.008 and a Hausdorff STD of 5.58. In the active selection setting, random selection again demonstrates higher variability, yielding a Dice STD of 0.05 and a Hausdorff STD of 21.28, whereas our combined uncertainty–diversity strategy substantially reduces variability, achieving a Dice STD of 0.0075 and a Hausdorff STD of 5.52.

\section{Discussion}
This paper presents a novel approach that combines cold-start selection with a subsequent active learning selection strategy. Across budgets and datasets, our cold-start clustering consistently matches or exceeds random seeding, often by a small but repeatable margin, and with lower variance across runs. In practice, this indicates that combining a simple medoid selection with a proportional farthest-point strategy provides more reliable early supervision than random sampling, improving both overlap- and boundary-based metrics, an important advantage under limited annotation budgets. Compared to the commonly used K-means-to-budget baseline, our initialization is generally competitive while avoiding the occasional severe failures observed with K-means when clusters are imbalanced or poorly separated. This effect is particularly pronounced for the Hausdorff distance, where our cold-start strategy consistently outperforms random selection across all datasets, whereas K-means-to-budget degrades performance in some cases.

After the cold start, the combined active learning acquisition that mixes uncertainty and diversity is generally as strong as, or slightly stronger than, either {uncertainty-only or diversity-only. Although the average gaps are modest, the hybrid tends to be more stable across rounds and budgets.

Our pipeline is applicable to both binary and multi-class segmentation tasks, indicating good portability across different problem settings. Performance depends on the quality of the feature space: stronger encoders typically improve clustering and distance-based selection and may further widen gaps over random. In addition, projecting the learned representations into a t-SNE space provides not only computational advantages for clustering and selection, but also improved interpretability by enabling intuitive visualization of the selected samples and their distribution in the feature space.

Our study has several limitations. First, computational constraints related to GPU memory and runtime precluded extensive experimentation with full 3D training; consequently, all experiments were conducted in a 2D setting, which may not fully reflect the behavior of the proposed method in three-dimensional scenarios. Second, during the active learning stage, we fixed the uncertainty–diversity trade-off to a single value ($\alpha=0.3$), rather than exploring a wider parameter sweep or dynamically adapting this balance across rounds or datasets. Finally, the effectiveness of the method depends on the quality of the underlying feature space, and we did not evaluate alternative representations, despite the fact that more expressive encoders typically improve clustering and distance-based selection and may further amplify performance gains over random sampling. Future work can address these topics.

\section{Conclusion}
This paper presents a novel approach that combines cold-start selection with a subsequent active learning selection strategy. The cold-start method includes structure-aware clustering encoder features, seeding with medoids and allocating the remaining budget via proportional farthest-first augmentation. The subsequent active selection method combines uncertainty and diversity criteria.

Results across three datasets spanning X-ray and MRI modalities and multiple anatomical targets show that our cold-start strategy consistently outperforms random seeding and exhibits lower run-to-run variability than the $k$-means-to-budget heuristic, which can occasionally fail in the presence of unbalanced clusters. In the active-learning phase, average gaps are smaller, yet the hybrid acquisition that mixes diversity and uncertainty remains consistently competitive with either signal alone across datasets and budgets. 

Notably, these performance gains are achieved with relatively low computational overhead by leveraging a pretrained encoder together with a lightweight t-SNE projection to guide sample selection, while also enhancing the explainability of the selection process. Taken together, the pipeline is simple, compute-efficient, and competitive compared to methods reported in the literature, providing a stable baseline for label-efficient medical image segmentation. 

\clearpage 
\newpage
\bibliographystyle{elsarticle-num}
\bibliography{res}    

\end{document}